\documentclass[conference]{IEEEtran}
\usepackage{cite}
\usepackage{ascmac}
\usepackage{algorithmic}
\usepackage[pdftex]{color}
\usepackage[whole]{bxcjkjatype}
\usepackage[pdftex]{graphicx}
\usepackage{caption}

\usepackage{url}
\usepackage{latexsym}

\begin{document}
\renewcommand{\algorithmicrequire}{\textbf{Input:}}
\renewcommand{\algorithmicensure}{\textbf{Output:}}
\renewcommand{\algorithmicforall}{\textbf{for each}}
%
\title{Discovery of Rare Causal Knowledge from Financial Statement Summaries}


\author{
\IEEEauthorblockN{Hiroki Sakaji}
\IEEEauthorblockA{The University of Tokyo,\\7-3-1 Hongo,\\
Bunkyo-ku, Tokyo 113-8654, Japan\\
Email: sakaji@sys.t.u-tokyo.ac.jp}\\
\IEEEauthorblockN{Jason Bennett}
\IEEEauthorblockA{Sumitomo Mitsui Asset Management \\
Company, Limited, 2-5-1 Atago, \\ Minato-ku, Tokyo 105-6228, Japan}\\
\and
\IEEEauthorblockN{Risa Murono}
\IEEEauthorblockA{Seikei University,\\3-3-1 Kichijoji-Kitamachi,\\
Musashino-shi, Tokyo 180-8633}\\ \\
\IEEEauthorblockN{Kiyoshi Izumi}
\IEEEauthorblockA{The University of Tokyo,\\7-3-1 Hongo,\\
Bunkyo-ku, Tokyo 113-8654, Japan}\\
\and
\IEEEauthorblockN{Hiroyuki Sakai}
\IEEEauthorblockA{Seikei University,\\3-3-1 Kichijoji-Kitamachi,\\
Musashino-shi, Tokyo 180-8633}\\ \\
}


%


\maketitle

\begin{abstract}
\renewcommand{\thefootnote}{\fnsymbol{footnote}}
\footnote[0]{© 2017 IEEE.  Personal use of this material is permitted.  Permission from IEEE must be obtained for all other uses, in any current or future media, including reprinting/republishing this material for advertising or promotional purposes, creating new collective works, for resale or redistribution to servers or lists, or reuse of any copyrighted component of this work in other works.}
\renewcommand{\thefootnote}{\arabic{footnote}}
What would happen if temperatures were subdued and result in a cool summer?
One can easily imagine that air conditioner, ice cream or beer sales would be suppressed as a result of this.
Less obvious is that agricultural shipments might be delayed, or that sound proofing material sales might decrease.
The ability to extract such causal knowledge is important, but it is also important to distinguish between cause-effect pairs that are known and those that are likely to be unknown, or rare.
Therefore, in this paper, we propose a method for extracting rare causal knowledge from Japanese financial statement summaries produced by companies.
Our method consists of three steps.
First, it extracts sentences that include causal knowledge from the summaries using a machine learning method based on an extended language ontology.
Second, it obtains causal knowledge from the extracted sentences using syntactic patterns.
Finally, it extracts the rarest causal knowledge from the knowledge it has obtained.
\end{abstract}



%
\IEEEpeerreviewmaketitle

\section{Introduction}
In recent years, artificial intelligence methods and technologies have been applied to various aspects of the financial market.
For example, a method of analyzing large amounts of financial information to support investment decisions has attracted much attention.
In addition, the number of individuals investing in securities markets is increasing, leading to an increasing need for technology to support the investment decisions of these individual investors.
For investors, information on corporate performance is important in making investment decisions, but as well as the company's performance itself, the causes and effects involved in that performance are important.
For example, individual investors could be informed that the level of ``demand for cooling'' could be increased by a ``hot summer'' by presenting them with information such as the cause ``hot summer'' and the effect ``increase in demand for cooling.''
Given the cause ``hot summer'' and the effect ``increase in demand for cooling,'' we can expect that companies involved in the cooling business will perform strongly in hot summers.
However, there are as many as 3,500 companies listed on the securities market, and in recent years, they have announced their financial results four times a year.
It takes a great deal of effort to manually compile the causes and effects involved in the performance of many companies.
We know, owing to causal knowledge, that the dollar will decline and the value of the yen will increase if US consumer prices rise significantly: if US prices rise, US products will be more expensive compared with Japanese products, resulting in US imports increasing and the dollar weakening against the yen.
If this causal knowledge could be gathered automatically, we might learn particularly important causal knowledge from it, leading to investment opportunities.
The ability to extract such causal knowledge is important, but it is also important to distinguish between cause-effect pairs that are known and those that are likely to be unknown, or rare.
Therefore, in this paper, we propose a method for extracting rare causal knowledge from Japanese financial statement summaries produced by companies.
This is important because merely listing all extracted causal knowledge does not provide sufficient support for individual investors.
For example, we believe that the above-mentioned cause ``hot summer'' and effect ``increase in demand for cooling'' are very general, highly likely to be common knowledge for individual investors and hence not useful information.
%

%
%

\section{Method overview}
This section presents an overview of our method.
In this research, we treat the combination of a cause expression and an effect expression as a piece of causal knowledge
For example, the sentence ``円高のため、日本経済は悪化した。({\it endaka no tame nihonkeizai ha akkashita}: Because of the yen's appreciation, the Japanese economy deteriorated.)'' includes cause expression ``円高({\it endaka}: the yen's appreciation)'' and effect expression ``日本経済は悪化した。({\it nihonkeizai ha akkashita}: the Japanese economy deteriorated.).''
Our method extracts these cause–effect expressions using clue expressions.
In this case, the clue expression is ``ため({\it tame}: because).''
Our method consists of the following three steps.
\begin{description}
  \item[Step 1] extracts sentences that include cause–effect expressions from Japanese financial statement summaries using a machine learning method.
  \item[Step 2] obtains cause–effect expressions from the extracted sentences using syntactic patterns.
  \item[Step 3] extracts the rarest causal knowledge from the knowledge obtained.
\end{description}

\section{Extraction of sentences that include cause-effect expressions}
\label{select}
In this section, we discuss how to extract sentences that include cause–effect expressions.
This is made more tricky by the fact that the clue expression ``ため({\it tame}: because)'' is important for extracting cause–effect expressions, but it can also be used to mean an objective.
For example, the sentence ``あなたのために、花を買った。({\it anata no {\bf tame} ni hana wo katta}: I bought some flowers {\bf for} you)'' includes the clue expression ``ため({\it tame}: because)'', but it does not have a causal meaning in this context.
We will therefore develop a method for extracting sentences that include cause and effect expressions that can cope with such situations.
Since this method uses a support vector machine (SVM) for extraction, we will now explain how to acquire features from PDF files of financial statement summaries.

\subsection{Acquring features}
To extract sentences that include cause and effect expressions, our method uses the features shown in \ref{flist}.
\begin{table}[h]
 \caption{List of features}
 \label{flist}
  \begin{center}
\begin{screen}
  {\bf Syntactic features}
  \begin{itemize}
  \item
    Pairs of particles
  \end{itemize}

  {\bf Semantic features}
  \begin{itemize}
  \item
    Extended language ontology
  \end{itemize}

  {\bf Other features}
  \begin{itemize}
  \item
    Part of speech of morphemes just before clue expressions
  \item
    Clue expressions
  \item
    Morpheme unigrams
  \item
    Morpheme bigrams
  \end{itemize}
\end{screen}
\end{center}
\end{table}

\begin{figure}[t]
 \begin{center}
  \includegraphics[width=\hsize]{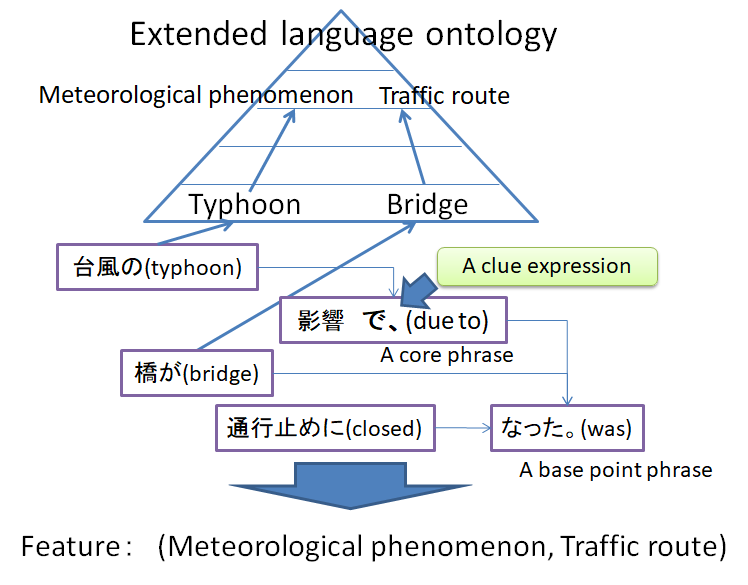}\\
  \caption{An exmaple of semantic feature}
  \label{ont}
 \end{center}
\end{figure}

We employ both syntactic and semantic features.
We aim to use expressions that are frequently used in cause and effect expressions in Japanese sentences as syntactic features.
For example, the sentence ``半導体{\bf の}需要回復{\bf を受けて}半導体メーカーが設備投資{\bf を}増やしている。({\it handoutai no jyuyoukaifuku woukete hanndoutaime-ka- ga setsubitoushi wo fuyashiteiru}: As semiconductor demand recovers, semiconductor manufacturers are increasing their capital investment.)'' has the following pattern of particles and clue expressions: ``... の ... を受けて ... を ... ({\it ... no ... woukete ... wo ...}).''
This pattern indicates that it is highly likely to be a cause–effect expression.
Our method therefore acquires particles that relate to clue expressions using a Japanese syntactic parser.
In addition, our method acquires words indicating causality using an extended language ontology \cite{kobayashi2010_en}.
Fig. \ref{ont} shows our method acquiring a semantic feature.
Here, a {\bf core phrase} is the last part of a phrase that includes a clue expression, and a {\bf base point phrase} is a phrase that is modified by the core phrase.
First, our method acquires words that modify the core phrase or the base point phrase.
Then, a tuple of concept words that is got by tracing extended language ontology using the words is acquired as a syntactic feature.
Here, each tuple consists of two concept words, one based on the core phrase and the other based on the base point phrase.

\section{Extracting cause-effect expressions}
\begin{figure}[t]
 \begin{center}
  \includegraphics[width=\hsize]{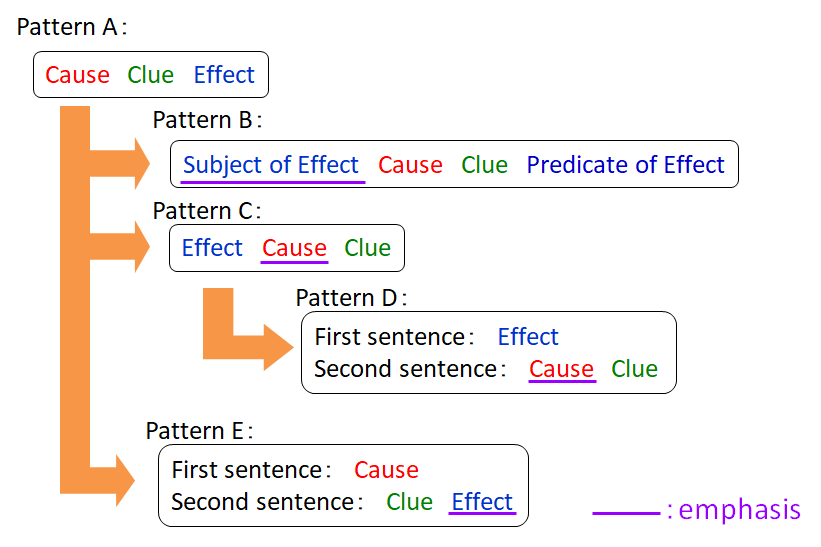}\\
  \caption{List of syntactic patterns}
  \label{patterns}
 \end{center}
\end{figure}

\begin{figure}[t]
 \begin{center}
  \includegraphics[width=\hsize]{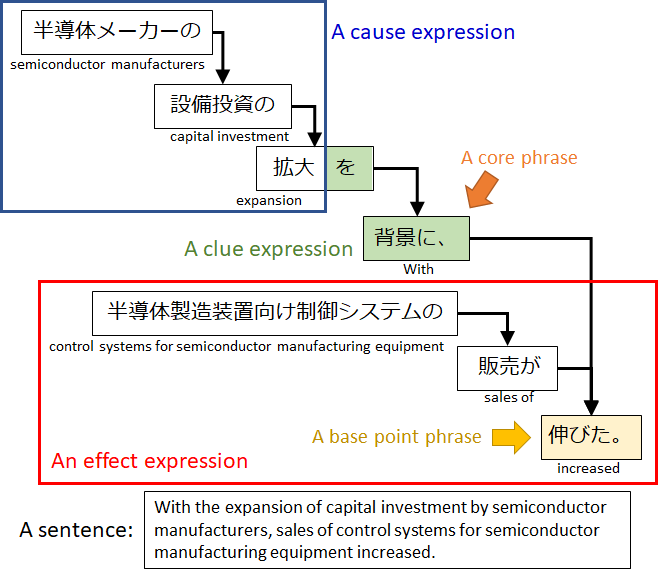}\\
  \caption{An example of pattern A}
  \label{pattern_a}
 \end{center}
\end{figure}

In this section, we explain how to extract cause–effect expressions from the sentences obtained in \ref{select}.
We employ a method by Sakaji et al.\cite{sakaji_pakm} to extract cause–effect expressions using four syntactic patterns.
They used a Japanese dependency analyzer \cite{cabocha_eng} to analyze Japanese syntax, but the recall of that method is too low for our needs.
However, the method is not enough about the recall.
Therefore, to improve the recall of the method, we add a new syntactic pattern (Pattern E); the extended list of syntactic patterns is shown in \ref{patterns}.
In Fig. \ref{patterns}, ``Cause'' indicates a cause expression, ``Effect'' indicates an effect expression and ``Clue'' indicates a clue expression.
We believe that Pattern A is the most basic pattern for expressing causality in Japanese, so the other patterns derive from Pattern A to emphasize either cause or effect, as expressed by the arrows in Fig. \ref{patterns}.
Emphasized expressions are underlined.
To illustrate the operation of our method, we now work through examples using two of the five syntactic patterns (Patterns A and C).
Fig. \ref{pattern_a} shows our method extracting cause and effect expressions using Pattern A.
It first identifies core and base point phrases using the clue expression ``を背景に({\it wohaikeini}: with).''
Then, the cause expression ``半導体メーカーの設備投資の拡大({\it handoutaime-ka-nosetsubitoushinokakudai}: expansion of capital investment by semiconductor manufacturers)'' is extracted by tracking back through the syntactic tree from the core phrase.
Finally, the effect expression ``半導体製造装置向け制御システムの販売が伸びた。({\it handoutaiseizousouchimukeseigyosisutemunohannbaiganobita}: sales of control systems for semiconductor manufacturing equipment increased)'' is extracted by tracking back through the syntactic tree from the base point phrase.

\begin{figure}[t]
 \begin{center}
  \includegraphics[width=\hsize]{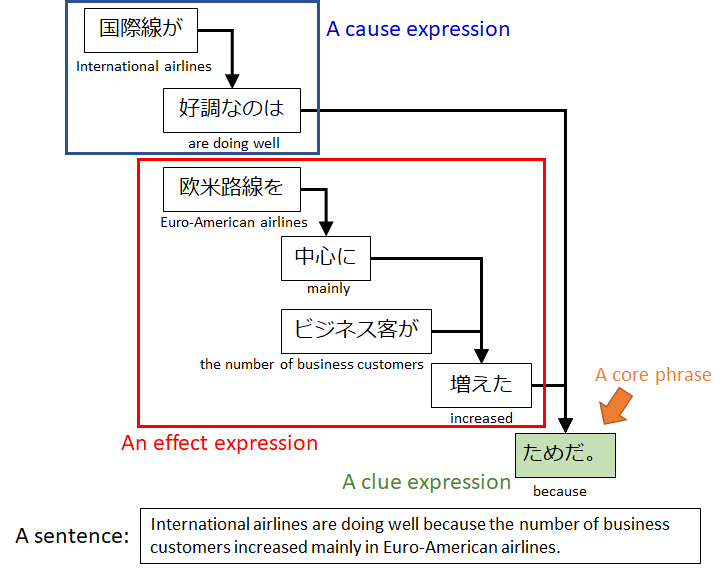}\\
  \caption{An example of pattern C}
  \label{pattern_c}
 \end{center}
\end{figure}

Fig. \ref{pattern_c} shows our method extracting cause and effect expressions using Pattern C.
It first identifies core and base point phrases using the clue expression ``ためだ。({\it tameda}: because).''
Then, the cause expression ``国際線が好調なのは({\it kokusaisengakoutyounanoha}: International airlines are doing well )'' is extracted by tracking back through the syntactic tree from the core phrase.
Finally, the effect expression ``欧米路線を中心にビジネス客が増えた({\it oubeirosenwotyushinnibijinesukyakugafueta}: the number of business customers increased mainly in Euro-American airlines)'' is extracted by tracking back through the syntactic tree from the base point phrase.

\section{Rare cause-effect expressions}
In this section, we introduce a method for extracting the rarest cause–effect expressions.
We first explain how to calculate the scores that are used to extract rare cause–effect expressions before discussing how to use these scores for the extraction process.

\subsection{Score calculation}
\label{scores}
Our method calculates scores (co-occurrence frequency and conditional probability) using the following steps.
\begin{description}
  \item[Step 1] acquires noun n-grams from cause expressions and effect expressions, which are called cause n-grams and effect n-grams, respectively.
  \item[Step 2] calculates the co-occurrence frequencies for all of the cause and effect n-grams.
  \item[Step 3] calculates the conditional probabilities of effect n-grams co-occurring with cause n-grams.
\end{description}

\subsection{Extracting rare cause–effect expressions}
Now we describe how to extract the rarest cause–effect expressions using the calculated scores with the algorithm shown in Fig. \ref{rare}.
\begin{figure}
\begin{algorithmic}[1]
\REQUIRE Keyword $w$ and Causal Information List $CI$ \\
$CI_i$ = (Cause Expression $c_i$, Effect Expression $e_i$, Company $cp_i$) \\
$c_i$: includes cause N-grams $\left(cn_{i0}, cn_{i1}, ..., cn_{in} \right)$ \\
$e_i$: includes effect N-grams $\left(en_{i0}, en_{i1}, ..., en_{im} \right)$
\ENSURE Rare Cause-Effect Expressions $RC$
\STATE $SC \leftarrow \emptyset$
\STATE $T \leftarrow \emptyset$

\FORALL{($c$, $e$, $cp$) $\in CI$}
\STATE $S_s \leftarrow 0$
\STATE $S_c \leftarrow 0$
\STATE $S_e \leftarrow 0$
\STATE $f_c \leftarrow 0$
\STATE $f_e \leftarrow 0$

\FORALL{$cn \in c$}
\IF{$cn \in$ CompanyKeywords($cp$)}
\STATE $S_c \leftarrow S_c$ + getCompanyKeywordScore($cn$)
\STATE $f_c \leftarrow f_c + 1$
\ENDIF
\ENDFOR

\FORALL{$en \in e$}
\STATE $S_s \leftarrow S_s$ + getConditionalProbability($w$, $en$)
\IF{$en \in$ CompanyKeywords($cp$)}
\STATE $S_e \leftarrow S_e$ + getCompanyKeywordScore($en$)
\STATE $f_e \leftarrow f_e + 1$
\ENDIF
\ENDFOR

\STATE $S_h \leftarrow 2 \left( f_c S_c f_e S_e \right) / \left( f_c S_c + f_e S_e \right)$
\STATE $T[(c, e, cp)] \leftarrow  S_h / S_s$
\ENDFOR

\STATE $RC \leftarrow$ getRareCausalKnowledge($T$)
\STATE \textbf{return} $RC$
\end{algorithmic}
\caption{Extraction of rare cause–effect expressions}
\label{rare}
\end{figure}
In Fig. \ref{rare}, the causal information list $CI$ consists of triplets formed from a cause expression, an effect expression and a company.
Here, the company is the company that created the document from which the cause and effect expressions were extracted, and $T$ is an associative array that uses triplets as keys and importance scores as values.
The methods CompanyKeywords($cp$), getCompanyKeywordScore($en$), getConditionalProbability($w$, $en$) and getRareCausalKnowledge($T$) are described in the following sections.

\subsubsection{CompanyKeywords}
CompanyKeywords($cp$) returns the list of keywords related to input company $cp$, and getCompanyKeywordScore($en$) returns the value of company keyword $en$.
The keywords returned by CompanyKeywords($cp$) are extracted using \ref{eq:company}.
\begin{equation}
    Score(w, cp) = \frac{W(w, S(cp))}{max_{w'}W(w', S(cp))},
    \label{eq:cs}
\end{equation}
\begin{equation}
    W(w, S(cp)) = TF(w, S(cp))H(w, S(cp))\log_2\frac{N}{df(w)},
    \label{eq:company}
\end{equation}
Here,
$S(cp)$ is the set of financial statement summary PDF files for company $cp$,
$TF(w, S(cp))$ is the frequency of word $w$ in $S(cp)$,
$df(w)$ is the number of PDF files that include word $w$,
$N$ is the number of companies used for this research,
$H(w, S(cp))$ is the entropy of word $e$ based on the probability $P(w, d)$ that word $e$ occurs in PDF $d$.
$P(w,d)$ and $H(w, S(cp))$ are calculated as follows:

\begin{equation}
    H(w, S(cp)) = - \sum_{d \in S(cp)} P(w,d)\log_2 P(w,d),
    \label{eq:ent}
\end{equation}
\begin{equation}
    P(w,d) = \frac{f(w,d)}{\sum_{d' \in S(cp)} f(w,d')},
    \label{eq:p}
\end{equation}
where
$f(w,d)$: is the number of word $w$ in pdf file $d$.
The scoring function $Score(w, cp)$ in \ref{eq:cs} assigns large values to words that appear frequently and evenly across the collection of PDF files for company $cp$, namely, $S(cp)$, and only appear in these specific PDF files.
The words with the highest $Score(w, cp)$ values are selected as company keywords.

\subsubsection{getConditionalProbability}
getConditionalProbability($w$, $en$) returns the conditional probability of the effect n-gram $en$ co-occurring with word $w$.
This conditional probability was calculated at Step 3 of Section \ref{scores}.

\subsubsection{getRareCausalKnowledge}
getRareCausalKnowledge($T$) extracts the $N$ cause–effect expressions with the highest scores as rare cause-effect expressions.
In this study, $N$ was 20.

\subsection{Algorithm description}
In Fig. \ref{rare}, line 22 represents the harmonic average the company keyword score, multiplied by the number of times it appears in either a cause or effect expression.
It is important to note that only in cases where a company keyword appears in both cause expressions and the effect expressions, will $S_h$ be non-zero.
In such cases, $S_h$ will be higher if the company keywords appearing in cause-effect expressions have larger keyword scores.
In line 23, we scale $S_h$ with $S_s$, the co-occurrence probability of $w$ with each n-gram occuring in the effect expression.
This ensures that when the co-occurence probability is low, i.e. the probability that a keyword occuring with n-grams in the effect expression is rare, $T[(c, e, cp)]$ has a high value.

\section{Evaluation}
\begin{figure}[h]
 \begin{center}
  \includegraphics[width=\hsize]{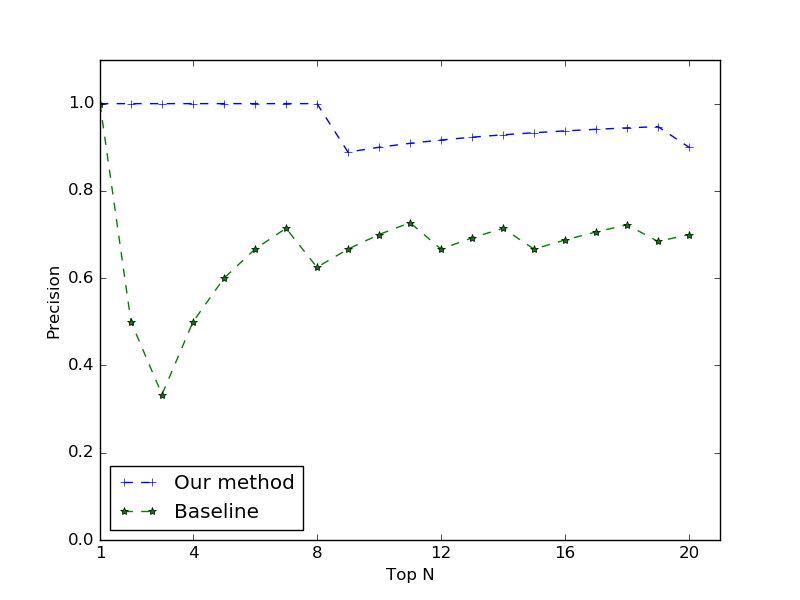}\\
  \caption{Precision results for the keyword ``Hot summer''}
  \label{hot-summer}
 \end{center}
\end{figure}

\begin{figure}[h]
 \begin{center}
  \includegraphics[width=\hsize]{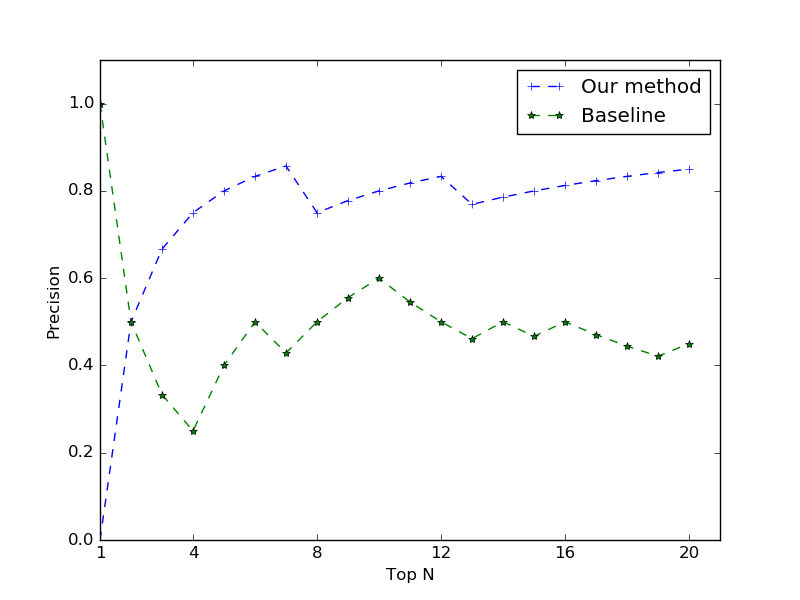}\\
  \caption{Precision results for the keyword ``Cold summer''}
  \label{cold-summer}
 \end{center}
\end{figure}

\begin{figure}[h]
 \begin{center}
  \includegraphics[width=\hsize]{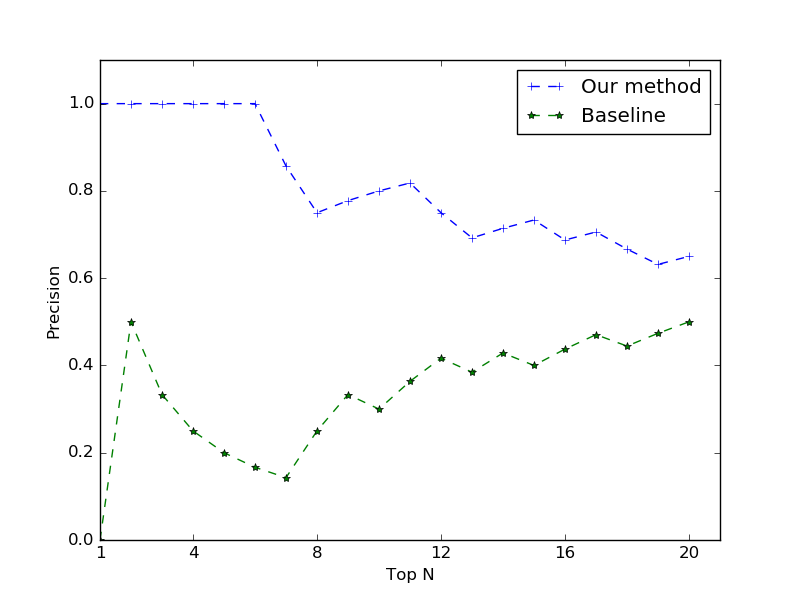}\\
  \caption{Precision results for the keyword ``Hot winter''}
  \label{hot-winter}
 \end{center}
\end{figure}

\begin{figure}[h]
 \begin{center}
  \includegraphics[width=\hsize]{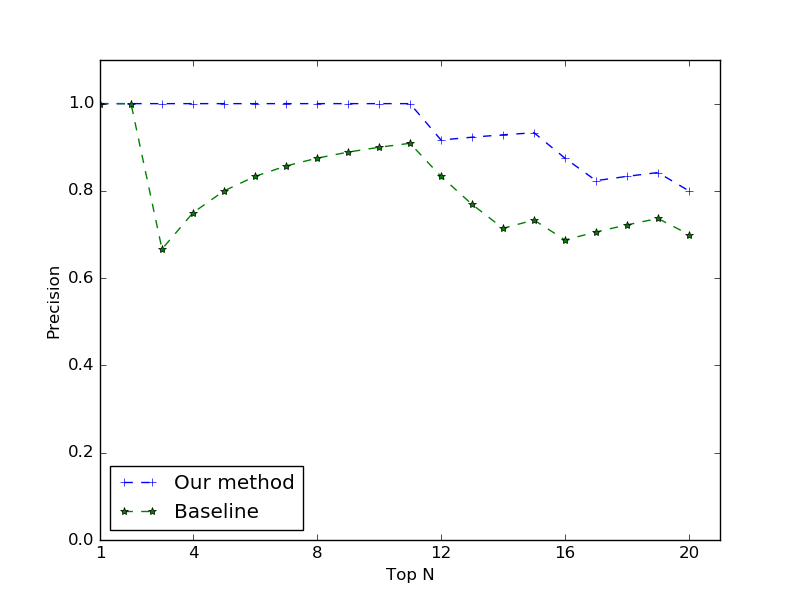}\\
  \caption{Precision results for the keyword ``Cold winter''}
  \label{cold-winter}
 \end{center}
\end{figure}

In this section, we evaluate the proposed method using 106,885 financial statement summaries in PDF form, which were acquired from 3,821 Japanese companies.
We used ``猛暑({\it mousho}: hot summer),'' ``冷夏({\it reika}: cold summer),'' ``暖冬({\it dantou}: hot winter)'' and ``厳冬({\it gentou}: cold winter)'' as Keywords.
And we used 1,000 sentences including clue expressions extracted randomly from the Nikkei newspaper articles of 1995 to 2005 as training data for extracting sentences that include cause-effect expressions.
A baseline was established by extracting rare cause–effect expressions based on the conditional probabilities calculated in Section \ref{scores}.
The expressions extracted by both methods were manually evaluated, and the results are shown in Table \ref{results}.
An individual investor with an investment history of 17 years tagged training data and test data.
Table \ref{extracted-rare} shows the rare cause–effect expressions extracted by our method.
\begin{table}[h]
\caption{Evaluation results}
\label{results}
\centering
\begin{tabular}{|l|r|r|}
\hline
& \multicolumn{2}{|c|}{Precision} \\ \cline{2-3}
& Our method & Baseline \\ \hline
Hot summer & 0.90 & 0.70 \\ \hline
Cold summer & 0.85 & 0.45 \\ \hline
Hot winter & 0.65 & 0.50 \\ \hline
Cold winter & 0.80 & 0.70 \\ \hline
Average & 0.80 & 0.59 \\
\hline
\end{tabular}
\end{table}
\begin{table*}[th]
\caption{Extracted rare cause–effect expressions}
\label{extracted-rare}
\centering
\begin{tabular}{|l|p{7cm}|p{7cm}|}
\hline
Keyword & Cause expression & Effect expression \\ \hline
Hot summer & 猛暑の影響 (Influence of a hot summer) & 除草関連用品、散水用品、日除け用品が好調に推移しました (Weed-related products, watering supplies and shading items were strong) \\ \hline
Cold summer & 冷夏の影響を受け、高冷地での開花が遅れた (Affected by a cold summer, flowering was delayed in high cold areas) & トルコギキョウは、お盆や秋のお彼岸の需要期には、品薄となり高値での取引となりました。 (In the high-demand periods of Bon and Higan, there was a shortage of Eustoma grandiflorum and it was sold at a high price) \\ \hline
Hot winter & 調理みそシーズン序盤の暖冬 (The beginning of the season when miso was often cooked was a hot winter) & ストレート鍋スープの出荷が伸び悩み (Shipments of potted soup are sluggish) \\ \hline
Cold winter & 猛暑と厳冬によりシニア層の来場が減った (Visits by senior citizens decreased due to a hot summer and a cold winter) & ゴルフ練習場においては、減収となりました。 (At the golf driving range, sales declined) \\
\hline
\end{tabular}
\end{table*}
We employed the mean average precision (MAP) as an evaluation metric \cite{rawashdeh2013folksonomy, krestel2009latent}, as follows:
\begin{equation}
    MAP = \frac{1}{|K|}\sum_{k=1}^{|K|} \left( \frac{1}{|N|} \sum_{n=1}^{|N|} T_n \times P@n \right)
    \label{eq:map}
\end{equation}
where
$K$ is the set of keywords,
$N$ is the number of extracted cause–effect expressions identified as rare,
$P@n$ is the precision at $n$ and
$T_n$ is a binary variable that is 1 if the cause–effect expression with rank $n$ is appropriate and 0 otherwise.
The MAP results are shown in Table \ref{map}.
\begin{table}[h]
\caption{MAP and average precision}
\label{map}
\centering
\begin{tabular}{|l|r|r|}
\hline
& MAP & Average Precision \\ \hline
Our method & 0.20 & 0.80 \\ \hline
Baseline & 0.15 & 0.59 \\
\hline
\end{tabular}
\end{table}
In order to verify the usefulness of the scoring function, graphs of the precision at $N$ were created for each keyword.
Figs. \ref{hot-summer}--\ref{cold-winter} show the results for the keywords ``hot summer,'' ``cold summer,'' ``hot winter'' and ``cold winter,'' respectively.

\section{Discussions}
Table II shows that our method achieved average precision 0.80.
On the other hand, baseline achieved average precision 0.59.
We consider that this result is enough for finding new companies, which have a chance to invest.
However, when the keyword was ``Hot winter,'' the precision of our method was 0.65.
Depending on the keyword, the accuracy of this method may be lower than 0.80, so additional experiments with other
keywords also need to be added.
The rare cause–effect expressions extracted by our method, shown in Table \ref{extracted-rare}, were interesting.
For example, the rare cause–effect expression extracted by our method for the keyword ``cold summer'' consisted of the cause expression ``Affected by a cold summer, flowering was delayed in high cold areas'' and the effect expression ``In the high-demand periods of Bon and Higan, there was a shortage of Eustoma grandiflorum and it was sold at a high price.''
Although it is easy to imagine that the flowering of flowers will be delayed by a cold summer, it is less obvious that the demand for flowers used for traditional Japanese events will increase owing to this flowering delay.
Likewise, the rare cause-effect expression extracted by our method for the keyword ``hot summer'' consisted of the cause expression ``Influence of a hot summer'' and the effect expression ``Weed-related products, watering supplies and shading items were strong.''
In this case, the cause–effect expression included the hidden causal knowledge that weeds grow well in hot summers, indicating the non-obvious conclusion that demand for weed-related products will increase in hot summers.
Figs. \ref{hot-summer}--\ref{cold-winter} show that our method was superior to the baseline overall, so we believe that the keywords produced by our method were effective for this task.

\subsection{Error analysis}
For analyzing our method, we checked errors of our method.
Our method extracted the below cause-effect expression as the rare cause-effect expression.
Table \ref{error1}, Table \ref{error2} and Table \ref{error3} show errors of our method.

\begin{table}[h]
\caption{An example of error for the keyword ``Cold winter''}
\label{error1}
\centering
\begin{tabular}{p{8cm}}
\hline \hline
{\bf Cause}: 販売効率の改善が進んだこと、また１１月以降気温が低下し例年になく厳冬となった (Improvement in sales efficiency advanced, and the temperature declined since November and it became the cold winter) \\
{\bf Effect}: 防寒衣料の販売が堅調に推移した (Sales of cold protection clothing remained steady) \\
\hline \hline
\end{tabular}
\end{table}

\begin{table}[h]
\caption{An example of error for the keyword ``Hot winter''}
\label{error2}
\centering
\begin{tabular}{p{8cm}}
\hline \hline
{\bf Cause}: 暖冬影響 (Hot winter effect) \\
{\bf Effect}: 住宅用空調機器の業界需要は、前期を下回りました (Industry demand for residential air conditioning equipment fell below the previous term) \\
\hline \hline
\end{tabular}
\end{table}

In Table \ref{error1}, since larger demand for cold protection clothing is common to cold winters, this cause-effect expression was judged to be unsuccessful.
In Table \ref{error2}, as the decline of demand for residential air conditioning equipment is also common to hot winters, this expression was also deemed unsuccessful.
To avoid extracting such expressions, it is necessary to improve scores and algorithms.

\begin{table}[h]
\caption{An example of error for the keyword ``Cold summer''}
\label{error3}
\centering
\begin{tabular}{p{8cm}}
\hline \hline
{\bf Cause}: 景気低迷と記録的な冷夏の影響 (The economic downturn and the impact of record cold summer) \\
{\bf Effect}: 飲料向け液糖を中心とした (Focusing on liquid sugar for beverages) \\
\hline \hline
\end{tabular}
\end{table}

The results in \ref{error3} show how our method extracted the inappropriate cause-effect expression, highlighting an error due to the Japanese dependency parser.
In order to avoid such errors, an algorithm which handles mistakes from the Japanese dependency parser is also necessary.

\section{Related work}
Much work has been done on the extraction of causal information from large corpuses.
Inui et al. proposed a method for acquiring causal relations ({\it cause}, {\it effect}, {\it precond} and {\it means}) from complex sentences containing the Japanese resultative connective ``ため({\it tame}: because)'' \cite{inui_en}, as this is a strong indicator of causal information.
However, their research used no other clue expressions, so their method cannot extract causal relations expressed by other clue expressions.
Khoo et al. proposed a method for extracting cause–effect information from newspaper articles by applying manually constructed patterns \cite{khoo}, as well as a method for extracting causal knowledge from medical databases by applying graphical patterns \cite{khoo2}.
However, in their research, both the cause and effect needed to occur together in the same sentence, so these methods are unable to extract causal knowledge spread over two sentences.
In contrast, our method can still extract causal knowledge in such cases.
Chang et al. proposed a method for extracting causal relationships between noun phrases using clue expressions and word pair probabilities \cite{chang}, defined as the probability that the pair forms a causal noun phrase.
They used a bootstrap method for teaching a naive Bayes causality classifier.
Girju proposed a method for automatic detection and extraction of causal relations based on clue phrases \cite{girju} where causal relations are expressed by pairs of noun phrases.
Girju used WordNet\cite{wordnet} to create semantic constraints for selecting candidate pairs, so her method cannot extract unknown phrases that are not in WordNet.
In contrast, our method can deal with causal knowledge expressed not only by noun phrases but also by verb phrases and sentences, and it can extract unknown expressions.
Sakai et al. proposed a method for extracting causal knowledge from Japanese financial articles concerning business performance \cite{sakai_ieice}, but their method only extracts cause expressions.
%
%
Bethard et al. proposed a method for classifying verb pairs that have causal relationships \cite{Bethard} using an SVM for classification.
Do et al. proposed a minimally supervised approach to identifying causal relationships between events in context  \cite{do2011minimally}.
Ittoo et al. proposed a method for extracting causal relationships from domain-specific texts \cite{ittoo2013minimally} by acquiring causal patterns from Wikipedia using Espresso \cite{espresso}.
Sadek et al. proposed a method for extracting Arabic causal relations using linguistic patterns \cite{sadek2016extracting} represented using regular expressions.
Low et al. proposed a semantic expectation-based knowledge extraction methodology (SEKE) for extracting causal relationships \cite{low2001semantic} that used WordNet as a thesaurus for extracting terms representing movement concepts.
In contrast, our method does not just extract cause–effect expressions but also identifies the rarest ones.

\section{Conclusions}
In this paper, we have proposed a method for extracting rare cause–effect expressions from Japanese financial statement summaries.
First, the method extracts sentences that include cause–effect expressions using machine learning.
We used extended language ontology as features of the support vector machine.
Second, it obtains cause–effect expressions from the extracted sentences using syntactic patterns.
The syntactic patterns consist of five patterns, and we added the new pattern E.
Finally, it extracts the rarest cause–effect expressions.
Using company-related keywords and an extraction algorithm we have developed, we were able to achieve an average precision of 0.8.
In the future, in order not to extract general cause-effect expressions, we have to improve scores and algorithms.
An algorithm which covers mistakes of Japanese dependency parser becomes necessary for avoiding Japanese parser errors.
Moreover, we will construct causal knowledge chains using the extracted cause–effect expressions as we believe that causal knowledge chains will help in finding investment opportunities.

\bibliographystyle{IEEEtran}
\bibliography{list}

\begin{thebibliography}{10}
\providecommand{\url}[1]{#1}
\csname url@samestyle\endcsname
\providecommand{\newblock}{\relax}
\providecommand{\bibinfo}[2]{#2}
\providecommand{\BIBentrySTDinterwordspacing}{\spaceskip=0pt\relax}
\providecommand{\BIBentryALTinterwordstretchfactor}{4}
\providecommand{\BIBentryALTinterwordspacing}{\spaceskip=\fontdimen2\font plus
\BIBentryALTinterwordstretchfactor\fontdimen3\font minus \fontdimen4\font\relax}
\providecommand{\BIBforeignlanguage}[2]{{%
\expandafter\ifx\csname l@#1\endcsname\relax
\typeout{** WARNING: IEEEtran.bst: No hyphenation pattern has been}%
\typeout{** loaded for the language `#1'. Using the pattern for}%
\typeout{** the default language instead.}%
\else
\language=\csname l@#1\endcsname
\fi
#2}}
\providecommand{\BIBdecl}{\relax}
\BIBdecl

\bibitem{kobayashi2010_en}
A.~Kobayashi, S.~Masuyama, and S.~Sekine, ``A method for automatic ontology construction using wikipedia,'' \emph{The IEICE transactions on information and systems (Japanese edition)}, vol. J93-D, no.~12, pp. 2597--2609, 2010.

\bibitem{sakaji_pakm}
H.~Sakaji, S.~Sekine, and S.~Masuyama, ``Extracting causal knowledge using clue phrases and syntactic patterns,'' in \emph{7th International Conference on Practical Aspects of Knowledge Management (PAKM)}, 2008, pp. 111--122.

\bibitem{cabocha_eng}
Y.~M. Taku~Kudo, ``Japanese dependency analysis using cascaded chunking,'' in \emph{CoNLL 2002: Proceedings of the 6th Conference on Natural Language Learning 2002 (COLING 2002 Post-Conference Workshops)}, 2002, pp. 63--69.

\bibitem{rawashdeh2013folksonomy}
M.~Rawashdeh, H.-N. Kim, J.~M. Alja’am, and A.~El~Saddik, ``Folksonomy link prediction based on a tripartite graph for tag recommendation,'' \emph{Journal of Intelligent Information Systems}, vol.~40, no.~2, pp. 307--325, 2013.

\bibitem{krestel2009latent}
R.~Krestel, P.~Fankhauser, and W.~Nejdl, ``Latent dirichlet allocation for tag recommendation,'' in \emph{Proceedings of the third ACM conference on Recommender systems}, 2009, pp. 61--68.

\bibitem{inui_en}
T.~Inui, K.~Inui, and Y.~Matsumoto, ``Acquiring causal knowledge from text using the connective marker {\it tame},'' \emph{Journal of Information Processing Society of Japan}, vol.~45, no.~3, pp. 919--933, 2004.

\bibitem{khoo}
C.~S. Khoo, J.~Kornfilt, R.~N. Oddy, and S.~H. Myaeng, ``Automatic extraction of cause-effect information from newspaper text without knowledge-based inferencing,'' \emph{Literary and Linguistic Computing}, vol.~13, no.~4, pp. 177--186, 1998.

\bibitem{khoo2}
C.~S. Khoo, S.~Chan, and Y.~Niu, ``Extracting causal knowledge from a medical database using graphical patterns,'' in \emph{Proceedings of the 38th ACL}, 2000, pp. 336--343.

\bibitem{chang}
D.-S. Chang and K.-S. Choi, ``Incremental cue phrase learning and bootstrapping method for causality extraction using cue phrase and word pair probabilities,'' \emph{Information Processing and Management}, vol.~42, no.~3, pp. 662--678, 2006.

\bibitem{girju}
R.~Girju, ``Automatic detection of causal relations for question answering,'' in \emph{In ACL Workshop on Multilingual Summarization and Question Answering}, 2003, pp. 76--83.

\bibitem{wordnet}
C.~Fellbaum, \emph{WordNet: An Electronic Lexical Database}.\hskip 1em plus 0.5em minus 0.4em\relax The MIT Press, 1998.

\bibitem{sakai_ieice}
H.~Sakai and S.~Masuyama, ``Cause information extraction from financial articles concerning business performance, ieice trans,'' \emph{IEICE Trans. Information and Systems}, vol. E91-D, no.~4, pp. 959--968, 2008.

\bibitem{Bethard}
S.~Bthard and J.~H.Martin, ``Learning semantic links from a corpus of parallel temporal and causal relations,'' in \emph{in Proceedings of ACL-08}, 2008, pp. 177--180.

\bibitem{do2011minimally}
Q.~X. Do, Y.~S. Chan, and D.~Roth, ``Minimally supervised event causality identification,'' in \emph{Proceedings of the Conference on Empirical Methods in Natural Language Processing}, 2011, pp. 294--303.

\bibitem{ittoo2013minimally}
A.~Ittoo and G.~Bouma, ``Minimally-supervised learning of domain-specific causal relations using an open-domain corpus as knowledge base,'' \emph{Data \& Knowledge Engineering}, vol.~88, pp. 142--163, 2013.

\bibitem{espresso}
P.~Pantel and M.~Pennacchiotti, ``Espresso: Leveraging generic patterns for automatically harvesting semantic relations,'' in \emph{in Proceedings of the 21st Internatinal Conference on Computational Linguistics and the 44th annual meeting of the ACL}, 2006, pp. 113--120.

\bibitem{sadek2016extracting}
J.~Sadek and F.~Meziane, ``Extracting arabic causal relations using linguistic patterns,'' \emph{ACM Transactions on Asian and Low-Resource Language Information Processing}, vol.~15, no.~3, p.~14, 2016.

\bibitem{low2001semantic}
B.-T. Low, K.~Chan, L.-L. Choi, M.-Y. Chin, and S.-L. Lay, ``Semantic expectation-based causation knowledge extraction: A study on hong kong stock movement analysis,'' in \emph{Pacific-Asia Conference on Knowledge Discovery and Data Mining (PAKDD)}, 2001, pp. 114--123.

\end{thebibliography}

\end{document}